\documentclass[oribibl]{llncs}
\usepackage{makeidx}  % allows for indexgeneration
\usepackage{bm}
\usepackage{algorithm}
\usepackage{algorithmicx}
\usepackage[noend]{algpseudocode}
\usepackage{tabularx}
\usepackage{graphicx}
\usepackage{subfigure}
\begin{document}
\frontmatter          % for the preliminaries
\pagestyle{headings}  % switches on printing of running heads

\title{Kaggle LSHTC4 Winning Solution}

\author{Antti Puurula\inst{1} \and Jesse Read\inst{2} \and Albert Bifet\inst{3}}
\institute{
Department of Computer Science, The University of Waikato, Private Bag 3105, Hamilton 3240, New Zealand
\and
Department of Information and Computer Science, Aalto University, FI-00076 Aalto, Espoo, Finland
\and
Huawei Noah's Ark Lab, Hong Kong Science Park, Shatin, Hong Kong, China
}

\maketitle              % typeset the title of the contribution

%Winning Model Documentation Template
%Competition winners are asked to provide their code and documentation. We've provided an example template below. Keep in mind that this document may be read %by people who have a machine learning background, have a software engineering background, or are entirely non-technical. It should be informative to each group.
%You may also be asked to participate in a Prize Winner Screencast or conference call with the competition sponsor.
%Related page: Model Submission Best Practices
%Winning Model Documentation
%Provide a word or PDF document on the winning model using the template below. The document should be well written and polished (in a suitable form to appear in a high impact scientific journal or conference). The document should be in English unless otherwise approved by Kaggle or the Competition Sponsor.
%Please describe the information for each section as applies to your model.
%Name:
%Location:
%Email:
%Competition:
%1. Summary
\section{Overview}
Our winning submission to the 2014 Kaggle competition for Large Scale Hierarchical Text Classification (LSHTC) consists mostly of an ensemble of sparse generative models extending Multinomial Naive Bayes. The base-classifiers consist of hierarchically smoothed models combining document, label, and hierarchy level Multinomials, with feature pre-processing using variants of TF-IDF and BM25. Additional diversification is introduced by different types of folds and random search optimization for different measures. The ensemble algorithm optimizes macroFscore by predicting the documents for each label, instead of the usual prediction of labels per document. Scores for documents are predicted by weighted voting of base-classifier outputs with a variant of Feature-Weighted Linear Stacking. The number of documents per label is chosen using label priors and thresholding of vote scores.\\

This document describes the models and software used to build our solution. Reproducing the results for our solution can be done by running the scripts included in the Kaggle package\footnote{https://kaggle2.blob.core.windows.net/competitions/kaggle/3634/media/\\LSHTC4\_winner\_solution.zip}. A package omitting precomputed result files is also distributed\footnote{https://kaggle2.blob.core.windows.net/competitions/kaggle/3634/media/\\LSHTC4\_winner\_solution\_omit\_resultsfiles.zip}. 
All code is open source, released under GNU GPL 2.0, and GPL 3.0 for Weka and Meka dependencies.\\

\section{Data Segmentation}

Source files: MAKE\_FILES, nfold\_sample\_corpus.py, fast\_sample\_corpus.py, shuffle\_data.py, count\_labelsets2.py\\

Training data segmentation is done by the script MAKE\_FILES, included in the code package. This segments the original training dataset train.txt by random 
sampling into portions for base-classifier training and for ensemble training. 2,341,782 documents are segmented for the former portion and 23,654 documents for 
the latter. The base-classifier training dataset dry\_train.txt is further sampled into 10
different folds, each with a 1000 document held-out portion dry\_dev.txt for parameter optimization. Folds 0-2 have exclusive and different sampled sets 
for dry\_dev.txt. Folds 3-5 sample dry\_train.txt randomly into 3 exlusive training subsets, with a shared optimization portion. Folds 6-9 segment dry\_train.txt in
the original data order into 4 exclusive training subsets, with a shared optimization portion. For all folds, the training datasets are further shuffled to improve 
the online pruning of parameters in training.\\

\section{Base-classifiers}

Source files: SGM-45l/, SGM-45l\_je/, Metaopt2.py, Make\_templates.py, results/, RUN\_DEVS, RUN\_EVALS, meka.jar\\

The base-classifiers consist mostly of sparse generative model extensions of Multinomial Naive Bayes (MNB). These extend MNB by introducing constrained finite 
mixtures at the document and hierarchy level nodes, and performing inference from the Multinomial node-conditional models using hierarchical smoothing, and  
kernel densities in case of document-conditional nodes. A special case is models using BM25 for kernel densities and no hierarchical smoothing. The models are 
stored in a sparse precomputed format, and inference using inverted indices is used to 
reduce the inference complexity according to the sparsity of the model. The constrained mixture modeling and sparse inference makes the models as 
scalable for text modeling as Naive Bayes and KNN, but with higher modelling accuracy. A detailed description of basic models of this type are given in 
\cite{Puurula:12, Puurula:13}. Since the LSHTC models can contain up to 100 million parameters for word counts, the models are provided as configuration
files in the package. Estimating the models from training data takes negligible time more compared to reading saved model files.\\ 

A development version of the SGMWeka toolkit\footnote{http://sourceforge.net/projects/sgmweka/} was customized to implement the models. The customized 
version is included as the Java source directory SGM-45l, and the program SGM\_Tests.java used for training and testing the models can be compiled without 
external dependencies. The documentation for SGMWeka version 1.4.4\footnote{http://sourceforge.net/p/sgmweka/wiki/SGMWeka\%20Documentation\%20v.1.4.4/} 
is accurate, but the development version contains additional
functionalities. A modified version is in the directory SGM-45l\_je. This includes the Meka toolkit\footnote{http://meka.sourceforge.net/} for doing 
multi-label decomposition used by one of the base-classifiers.\\

The script Metaopt2.py optimizes a base-classifier on a development set according to a chosen performance measure, by iteratively estimating the classifier 
and classifying the development data portion. The script RUN\_DEVS runs the development and compresses the log files. The configuration files for Metaopt2.py
describes all the parameters provided to a SGM\_Tests call, as well as the optimization measure to extract from the last line of the SGM\_Tests log file. 
Metaopt2.py performs a Gaussian Random Search \cite{Puurula:12c} for the chosen parameters, constrained and transformed according to the configuration file.
The directories results\_* contain the first and last parameter configuration file for each base-classifier type, after a 40x8 iteration random search. Some classifiers
were constructed by copying the parameters for similar folds (3,4,5), and some used manually chosen parameter configurations. These classifiers have the
final iteration parameter file wikip\_large\_X\_params.txt\_39\_0, but not the initial file wikip\_large\_X\_params.txt. The script Make\_templates.py makes 
the parameter template files as specified in the global variable "configs".\\

The template files describe the model by suffixing the file name with modifications. For example, 
"mnb\_mafs2\_s8\_lp\_u\_jm2\_bm18ti\_pct0\_ps5\_thr16.template" modifies a Multinomial Naive Bayes by optimizing the parameters for a modified version of 
macro-Fscore (\_mafs2), uses data fold 8 (\_s8), the Label Powerset method for multi-label classification (\_lp), smoothing by a uniform background distribution 
(\_u), a BM25 variant for feature weighting (\_bm18ti), uses a safe pruning of pre-computed parameters (\_pct0), constrains the scaling of label prior (\_ps5) and
uses 16 threads for parallel classification.\\

\begin{table}
\begin{tabular}{ l | c | l | l }
\hline
id & excluded & parameter configuration & maFscore \\
\hline
7 &  & mafs3\_s1\_uc1\_jm3\_bm18ti\_pci7\_pct0\_psX\_fb\_iw2 & 0.4155\\
9 &  & mafs3\_s1\_uc1\_jm3\_bm18ti\_pci7\_pct0\_psX\_iw2 & 0.4082\\
11 & X & mafs3\_s2\_uc1\_jm2\_bm18tid\_pci7\_pct0\_ps8\_iw1 & 0.3993\\
13 &  & mafs3\_s3\_kd\_u\_jm3\_kdp5\_bm18ti\_pct0\_ps7\_iw2 & 0.3982\\
17 &  & mafs3\_s4\_kd\_u\_jm3\_kdp5\_bm18ti\_pct0\_ps7\_iw2 & 0.3982\\
8 &  & mafs3\_s1\_uc1\_jm3\_bm18ti\_pci7\_pct0\_psX\_iw1 & 0.3866\\
10 & X & mafs3\_s2\_u\_lp\_jm2\_bm18tib\_pct0\_ps7\_iw0 & 0.3795\\
20 &  & mafs3\_s5\_kd\_u\_jm3\_kdp5\_bm18ti\_pct0\_ps7\_iw0 & 0.3771\\
12 &  & mafs3\_s3\_kd\_u\_jm3\_kdp5\_bm18ti\_pct0\_ps7\_iw0 & 0.3763\\
6 &  & mafs3\_s1\_u\_jm3\_bm18ti\_pct0\_ps7\_iw0 & 0.3689\\
16 &  & mafs3\_s4\_kd\_u\_jm3\_kdp5\_bm18ti\_pct0\_ps7\_iw0 & 0.3615\\
14 &  & mafs3\_s3\_kd\_uc1\_jm2\_kdp5\_bm18tid\_pct0\_ps8\_iw1 & 0.3466\\
5 &  & mafs3\_s0\_kd\_nobo\_bm25c2\_mi2\_ps2\_iw0 & 0.3380\\
19 &  & mafs3\_s4\_u\_jm2\_bm18tib\_pci6\_pct0\_ps7\_cs0\_iw2 & 0.3346\\
18 & X & mafs3\_s4\_u\_jm2\_bm18tib\_mc0\_pci6\_pct0\_ps7\_cs0\_iw0 & 0.3114\\
21 &  & mafs3\_s5\_u\_jm2\_bm18tib\_mc0\_pci6\_pct0\_ps7\_cs0\_iw0 & 0.3091\\
15 &  & mafs3\_s3\_u\_jm2\_bm18tib\_mc0\_pci6\_pct0\_ps7\_cs0\_iw0 & 0.3082\\
33 &  & mafs\_s2\_lp\_u\_jm5\_pd2\_bm16ti\_mc0\_pct0\_ps0 & 0.2860\\
50 &  & mjac\_s2\_kd\_nobo\_bm25c2\_mc0\_mlc0\_ps2\_lt5\_mr0\_tk1 & 0.2856\\
0 &  & mafs2\_s2\_lp\_u\_jm2\_bm18tib\_mc0\_pct0\_ps5 & 0.2815\\
28 & X & mafs\_s1\_kd\_nobo\_bm25c2\_mc0\_mlc0\_ps2\_lt5\_mr0\_tk2 & 0.2805\\
32 &  & mafs\_s2\_lp\_u\_jm4\_bm20ti\_mc0\_pct0\_ps2 & 0.2723\\
27 &  & mafs\_s0\_lp\_u\_jm2\_bm18tic\_fb3\_mc0\_pct0\_ps6 & 0.2686\\
44 & X & mjac\_s0\_lp\_u\_jm2\_pd2\_tXiX3\_fb2\_mc0\_pci1\_pct0\_ps0 & 0.2678\\
52 &  & ndcg5b\_s4\_kd\_u\_jm2\_kdp5\_bm18tib\_mc0\_pci0\_pct0\_mlc0\_ps6\_tk0 & 0.2659\\
51 &  & ndcg5b\_s3\_kd\_u\_jm2\_kdp5\_bm18tib\_mc0\_pci0\_pct0\_mlc0\_ps6\_tk0 & 0.2650\\
53 &  & ndcg5b\_s5\_kd\_u\_jm2\_kdp5\_bm18tib\_mc0\_pci0\_pct0\_mlc0\_ps6\_tk0 & 0.2643\\
30 &  & mafs\_s1\_lp\_u\_jm6\_tiX5\_mc0\_pct0\_ps0 & 0.2618\\
29 & X & mafs\_s1\_lp\_u\_jm4\_pd2\_tXiX2\_fb2\_mc0\_pct0\_ps0 & 0.2612\\
45 & X & mjac\_s0\_lp\_u\_jm2\_tiX3\_mc0\_pct0\_ps0 & 0.2592\\
31 & X & mafs\_s2\_lp\_u\_jm4\_bm18ti\_mc0\_pct0\_ps2 & 0.2567\\
23 &  & mafs3\_s7\_kd\_uc1\_jm2\_kdp5\_bm18tid\_mc0\_pci1\_pct0\_ps8\_iw1\_ch80 & 0.2550\\
42 &  & mifs\_s2\_lp\_u\_jm2\_bm18tib\_fb3\_mc0\_pct0\_ps5 & 0.2530\\
46 & X & mjac\_s0\_lp\_u\_jm4\_bm15ti\_mc0\_pct0\_ps0 & 0.2489\\
22 &  & mafs3\_s6\_kd\_uc1\_jm2\_kdp5\_bm18tid\_mc0\_pci1\_pct0\_ps8\_iw1\_ch80 & 0.2444\\
35 &  & mafs\_s4\_kd\_u\_jm3\_kdp5\_tXiX2\_mc0\_pci0\_pct0\_mlc0\_ps5\_lt5\_mr0\_tk2 & 0.2441\\
34 &  & mafs\_s3\_kd\_u\_jm3\_kdp5\_tXiX2\_mc0\_pci0\_pct0\_mlc0\_ps5\_lt5\_mr0\_tk2 & 0.2422\\
36 &  & mafs\_s5\_kd\_u\_jm3\_kdp5\_tXiX2\_mc0\_pci0\_pct0\_mlc0\_ps5\_lt5\_mr0\_tk2 & 0.2421\\
49 &  & mjac\_s1\_u\_jm3\_tiX1\_mc0\_pci1\_pct0\_mlc0\_ps1\_lt1\_mr0 & 0.2410\\
48 & X & mjac\_s1\_u\_jm2\_tiX1\_mc0\_pct0\_mlc0\_ps2\_lt2\_mr0\_tk0 & 0.2395\\
24 &  & mafs3\_s8\_kd\_uc1\_jm2\_kdp5\_bm18tid\_mc0\_pci1\_pct0\_ps8\_iw1\_ch80 & 0.2335\\
26 & X & mafs\_s0\_lp\_u\_jm2\_bm18tib\_mc0\_pct0\_ps5 & 0.2245\\
41 &  & mifs\_s1\_lp\_u\_jm2\_bm18tib\_mc0\_pct0\_ps5 & 0.2232\\
25 &  & mafs3\_s9\_kd\_uc1\_jm2\_kdp5\_bm18tid\_mc0\_pci1\_pct0\_ps8\_iw1\_ch80 & 0.2108\\
47 &  & mjac\_s0\_u\_jm3\_bm18ti\_pct0\_ps5\_je & 0.2040\\
43 &  & mjac\_s0\_lp\_bm25c1\_mc0\_mlc0\_ps3 & 0.1924\\
38 &  & mafs\_s7\_kd\_u\_jm3\_kdp1\_bm18ti\_mc0\_pci1\_pct0\_ps5\_lt5\_mr1\_tk2\_ch80 & 0.1787\\
39 &  & mafs\_s8\_kd\_u\_jm3\_kdp1\_bm18ti\_mc0\_pci1\_pct0\_ps5\_lt5\_mr1\_tk2\_ch80 & 0.1632\\
2 &  & mafs2\_s7\_lp\_u\_jm2\_bm18ti\_pct0\_ps5 & 0.1554\\
1 & X & mafs2\_s6\_lp\_u\_jm2\_bm18ti\_pct0\_ps5 & 0.1529\\
3 &  & mafs2\_s8\_lp\_u\_jm2\_bm18ti\_pct0\_ps5 & 0.1513\\
37 &  & mafs\_s6\_kd\_u\_jm3\_kdp1\_bm18ti\_mc0\_pci1\_pct0\_ps5\_lt5\_mr1\_tk2\_ch80 & 0.1469\\
40 &  & mafs\_s9\_kd\_u\_jm3\_kdp1\_bm18ti\_mc0\_pci1\_pct0\_ps5\_lt5\_mr1\_tk2\_ch80 & 0.1452\\
4 &  & mafs2\_s9\_lp\_u\_jm2\_bm18ti\_pct0\_ps5 & 0.1357\\
\hline
\end{tabular}
\caption{Base-classifiers sorted in the order of comb\_dev.txt macro-averaged Fscore, computed over the labels occurring in the set. 
Excluded models were removed by model selection from the ensemble.}
\label{table1}
\end{table}

Some of the modifications have little influence on the results, such as \_thr16 that instructs SGM\_Tests to use 16 threads. More detailed 
explanations of the 
important modifications are given in the following sections. A total of 54 base-classifiers are used in the ensemble, selected down to 42 
base-classifiers by model selection. Table \ref{table1} shows the base-classifiers sorted according to comb\_dev.txt macro-averaged Fscore.
It should be noted that the parameter ranges for some of the modifications were adjusted during the competition, and the parameter ranges 
in the individual template files can differ from those in Make\_templates.py.\\

The word count vectors for LSHTC were preprocessed by the organizers to remove common words, stopwords 
and short words, as can be seen from looking at the distributions of words in the vectors. This causes problems for some models such as 
Multinomial models of text, that assume word vectors to distribute normally. Feature transforms and weighting can be used to correct this. 
Feature weighting is done by each base-classifier separately, using variants of TF-IDF and BM25. All models use 1-3 parameters to optimize 
the feature weighting on the dry\_dev.txt portion of the fold. A variant of BM25 that proved most successful has the suffix "\_bm18ti". As 
seen in TFIDF.java, this combines the term count normalization of BM25 with the parameterized length normalization and idf weighting from 
TF-IDF that has been used earlier \cite{Puurula:12c}.\\

The Multinomials use hierarchical smoothing with a uniform background distribution \cite{Puurula:13}. The variant "\_uc1"
uses a uniform distribution interpolated with a collection model, improving the accuracy by a small amount. All models use Jelinek-Mercer "\_jmX" 
for smoothing label and hierarchy level Multinomials, and Dirichlet Prior smoothing "\_kdpX" for smoothing kernel density document models. The 
feature selection done by the organizers cause very unusual smoothing parameter configurations to be optimal. With Jelinek-Mercer values less 
than a heavy amount such as 0.98 become rapidly worse, with some models using a smoothing coefficient of 0.999.\\

Parameter pruning is chosen by the modifiers "\_mcX", "\_pciX", "\_pctX", "\_mlcX". "\_mcX" prunes word features based on their frequency.
"\_pciX" selects on-line pruning of conditional parameters, "\_pctX" performs mostly safe pruning of precomputed conditional parameters, 
"\_mlcX" prunes labels based on their frequency.\\

One special classifier is the variant using the modifer "\_je". This requires a development version of the Meka toolkit and the other files in the 
directory /SGM-45l\_je. This model does classification with label powersets decomposed into meta-labels, and transforms the meta-labels back 
into labelsets after classification. The labelset decompositions are stored in a precomputed file loaded by the modified version of SGM\_Tests.\\

Kernel densities are selected with the modifier "\_kd", passing -kernel\_densities to SGM\_Tests. This constructs document-conditional models,
and computes label-conditional probabilities using the document-conditionals as kernel densities \cite{Puurula:13}. The modifiers "\_csX" load
the LSHTC4 label hierarchy, and use random parent nodes to smooth the label-conditional Multinomials. The Label Powerset 
method for mapping a multi-label problem into a multi-class problem is done by the modifier "\_lp", passing -label\_powerset to SGM\_Tests.\\

The modifier "\_nobo" combined with "\_kd" produces models for document instances with no back-off smoothing by label-conditional models.
The modifiers "\_bm25X" use BM25 instead of Multinomial distances. Combined with "\_kd" and "\_nobo", this produces a model that uses BM25
for kernel densities of each label.\\

The modifiers "\_ndcg5",  "\_mjac", "\_mifs" and "\_mafsX" choose the optimization measure for MetaOpt2.py. These correspond to NDCG@5,
Mean of Jaccard scores per instance, micro-averaged Fscore, macro-averaged Fscore and surrogate measures for maFscore. It was noticed
early in the competition that computing and optimizing maFscore is problematic, since not all labels are present in the training set, and any
subset chosen for optimization will contain only a tiny fraction of the 325k+ labels, with the rest being missing labels. Since most labels are missing
labels, and any number of false positives for a missing label will equal an fscore of 0, optimizing maFscore becomes problematic. The "\_mafsX"
surrogates used two attempts to penalize for false positives of missing labels, but these was abandoned for a method that allows optimizing
macroFscore better without producing too many instances per label.\\

The modifiers "\_iwX" select a method developed in this competition. This causes the base-classifier to predict instances per label, instead of
labels per instance. A sorted list of the best scores for each label is stored, and for each classified instance the lists for labels are updated. 
A full distribution of labels is computed for each instance, and the label$\rightarrow$instance scores are computed from the rank of the label 
for each instance. After classification of the dataset, the sparse label$\rightarrow$instances scores are transposed and outputted and evaluated 
in the instance$\rightarrow$labels format. The arguments 
-instantiate\_weight X and -instantiate\_threshold X passed to SGM\_Tests control the number of top scoring instances stored for each label. The
ensemble combination uses transposed prediction of the same kind to do the classification.\\
 
\section{Ensemble Model}

Source files: RUN\_METACOMB, MetaComb2.java, TransposeFile.py, SelectClassifiers.py, SelectDevLabels.py, comb\_dev\_results/, eval\_results/, weka.jar\\

The ensemble model is built on our earlier LSHTC3 ensemble \cite{Puurula:12c}, but performs classification by predicting instances per label. 
The classifier vote weight prediction is a case of Feature-Weighted Linear Stacking \cite{Sill:09}, but the regression models are trained separately
for each base-classifier, using reference weights that approximate optimal weights per label in a development set.\\

The base-classifier result files are tranposed from a document$\rightarrow$labels per line format to a label$\rightarrow$documents per line format. 
After prediction the
ensemble result file is transposed back to the document$\rightarrow$labels per line .csv format used by the competition. The script RUN\_METACOMB performs
all the required steps, using the result files stored in /comb\_dev\_results for training the ensemble and /eval\_results to do the classifier combination.\\

Metacomb2.java perfoms the ensemble classification. The ensemble uses linear regression models to predict the weight of each base-classifier, using 
metafeatures computed from label information and
classifier outputs to predict the optimal classifier weight for each label. The most useful metafeatures in the LSHTC3 submission used labelset correlation
features between the base-classifiers for each document instance \cite{Puurula:12c}. This ensemble uses instance-set correlation features for each label 
analogously.\\

\begin{table}
\begin{tabular}{ l | l }
\hline
metafeature & description \\
\hline
labelProb & indicator feature for low-frequency labels ($<$10) \\
labelProb2 & indicator feature for high-frequency labels ($>$50) \\
uniqInstancesets & \# different instance sets in the classifier outputs \\
maxVotes & \# votes given to most voted instance set \\
minInstFreq\_i & the frequency of least frequent instance in the output of classifier $i$\\
maxInstFreq\_i & the frequency of most frequent instance in the output of classifier $i$\\ 
minInstCount\_i & the count of the lowest count instance in the output of classifier $i$\\
instCount\_i & \# instances in the output of classifier $i$\\
emptySet\_i & indicator if the classifier output $i$ has no instances for the label\\
setCount\_i & \# of classifiers with the same output as classifier $i$\\
modePrec\_i & precision of classifier $i$ using the mode of outputs as reference\\
modeRec\_i & recall of classifier $i$ using the mode of outputs as reference\\
modeJaccard\_i & Jaccard similarity of classifier $i$ and the mode of outputs\\
maxPrec\_i\_j & intersection of classifier $i$ and $j$ outputs, divided by maximum length\\
\hline
\end{tabular}
\caption{Metafeatures used for voting classifier weights. Metafeatures are computed for each label, given the instance set outputs from each base-classifier. 
Regression models for each baseclassifier weight uses the features that match the classifier id $i$, and not the metafeatures for other classifiers. Metafeature
maxPrec\_i\_j is computed for all the other base-classifiers $j$, resulting in 42-1 additional metafeatures. Metafeatures are normalized and log-transformed
based on development set performance.}
\label{table2}
\end{table}

Table \ref{table2} shows the metafeatures used by MetaComb2.java. For efficiency and memory use, MetaComb2 adds the correlation metafeatures
to each base-classifier before predicting the vote weights, and doesn't keep all possible metafeatures in memory at any time. This keeps the memory 
complexity of the ensemble combination linear in the number of base-classifiers. Functions constuctData(), pruneGlobalFeatures() and addLocalFeatures()
in MetaComb2.java show how the features are constructed as Weka \cite{Hall:09} Instances.\\

\begin{figure}
\centering
{\includegraphics[scale=0.5, trim=20 20 80 50, clip=true]{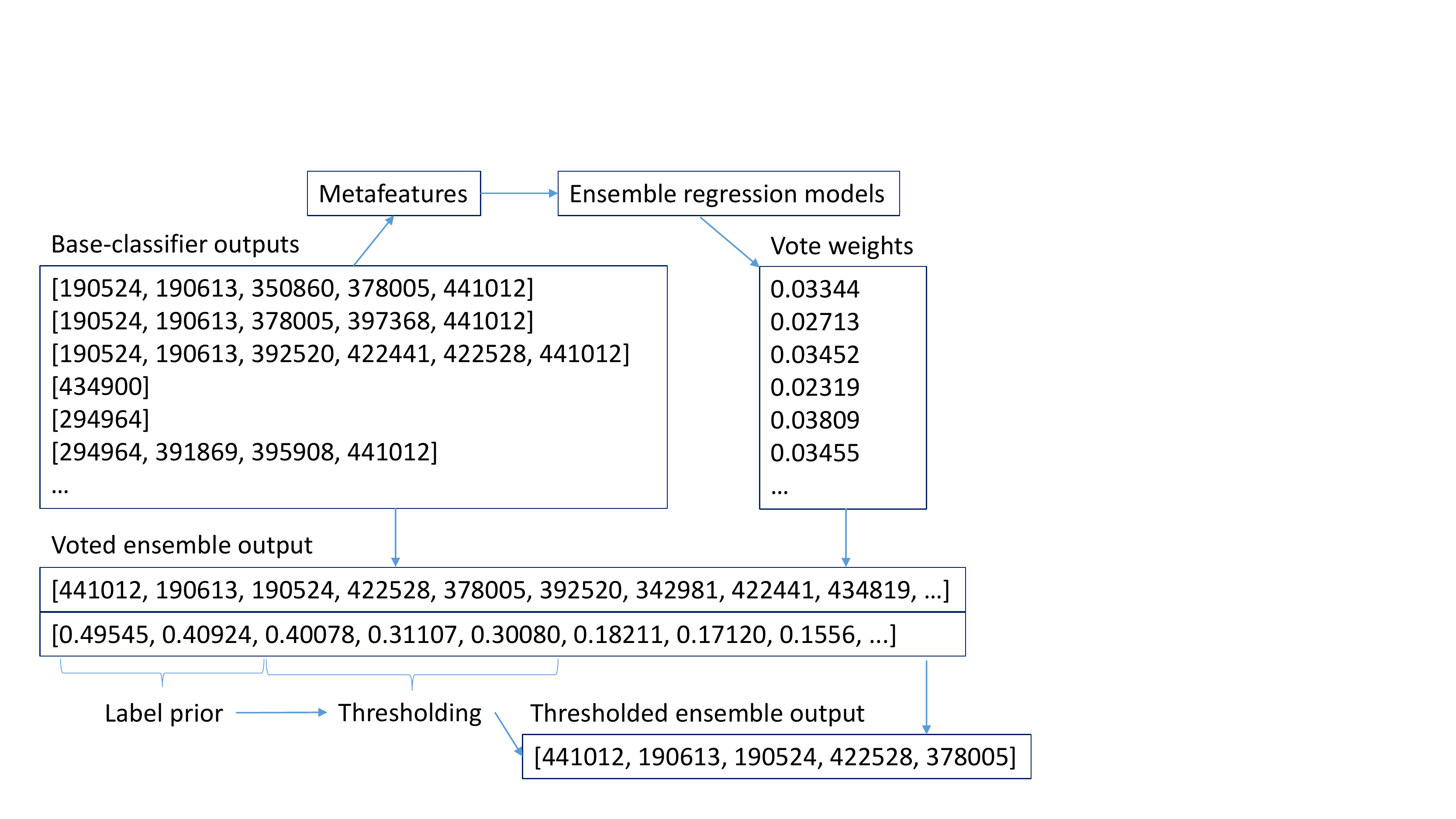}}
\caption{Ensemble voting and selection of instances from the base-classifier outputs.}
\label{fig2}
\end{figure}

The regression models use Weka LinearRegression for implementing the variant of Feature-Weighted Linear Stacking. For each label in comb\_dev.txt, 
optimal reference weights are approximated by distributing a weight of 1 uniformly to the base-classifiers that score highest on the performance measure. Initially
fscore was used as the measure, as averaging the fscores across the labels gives maFscore. This however doesn't use rank information in the instance sets. 
A small improvement in maFscore was gained by using a similar measure that takes rank information into account. approximateOracleWeights() and 
updateEvaluationResults() in MetaComb2.java show how the reference vote weights are constructed.\\

Following vote weight prediction, the label$\rightarrow$instances scores are summed for each instance from the weighted votes in the function voteFold(). 
A combination of label prior information and thresholding similar to one used in the base-classifiers is used to choose the number of instances per label. 
The label prior information selects a number of instances for the label proportional to the label frequency in training data, multiplied by the parameter $0.95$
passed to set\_instantiate(). The thresholding then includes to the set all instances with score more than $0.5$ of the mean of the initial instance set scores. 
Figure \ref{fig2} illustrates the ensemble combination and selection of instances.\\

Development of the ensemble by n-fold cross-validation can be done by changing the global variable "developmentRun" in MetaComb2.java to 1. Selection of 
base-classifiers can be done by giving the classifiers to remove as integer arguments to MetaComb2. The list of removed classifiers used in the final
evaluation run in RUN\_METACOMB was developed by running the classifier selection script SelectClassifiers.py with the n-fold crossvalidation. 
SelectClassifiers.py performs hill-climbing searches, maximizing the output of MetaComb2 by removing and adding classifiers to the ensemble.\\

\section{How to Generate the Solution}
The programs and scripts described above can be run to produce the winning submission file. Some of the programs can take considerable computing resources
to produce. Both optimizing the base-classifier parameters and classifying the 452k document test set can take several days or more, depending on the model. 
We used a handful of quadcore i7-2600 CPU processors with 16GB RAM over the competition period to develop and optimize the models. At least 16GB RAM is 
required to store the word counts reaching 100M parameters. Ensemble combination takes less than 8GB memory, and can be computed from the provided 
.results files. The base-classifier result files are included in the distribution, as computing these takes considerable time.\\

For optimizing base-classifiers, compile SGM\_Tests.java with javac, configure Make\_templates.py or copy an existing template, and run RUN\_DEVS. For classifying the 
comb\_dev.txt and test.txt results with a base-classifier, configure and run RUN\_EVALS. For combining the base-classifier results with the ensemble, run RUN\_METACOMB. 
The global variables in each script can be modified to change configurations.\\

\section{What Gave us the First Place?}

The competition posed a number of complications different from usual Kaggle competitions. Most of our tools were developed over the last LSHTC challenges, and this
gave us a big advantage. The biggest complication in the competition was scalability of both the base-classifiers and ensemble. Our solution uses sparse storage and 
inverted indices for inference, a modeling idea that enabled us to use an ensemble of tens of base-classifiers. With the SGMWeka toolkit we could combine 
parameterized feature weighting \cite{Puurula:12c}, hierarchical smoothing \cite{Puurula:13}, kernel densities \cite{Puurula:13}, model-based feedback \cite{Puurula:13b}, etc. Other participants used KNN with inverted indices, but our 
solution provides a diversity of structured probabilistic base-classifiers with much better modeling accuracies.\\

Another complication was the preprocessed pruned feature vectors. This made usual Multinomial or Language Model solutions usable only with very untypical 
and heavy use of linear interpolation smoothing. The commonly used TF-IDF feature transforms also corrected the problem only somewhat. Our solutions for 
smoothing and feature weighting with a customized BM25 variant took extensive experimentation to discover, but improved the accuracy considerably. It is likely 
that the other teams had less sophisticated text similarity measures available, and the ones having good measures scored better in the contest.\\

\begin{figure}
\centering
{\includegraphics[scale=0.7, trim=245 130 230 120, clip=true]{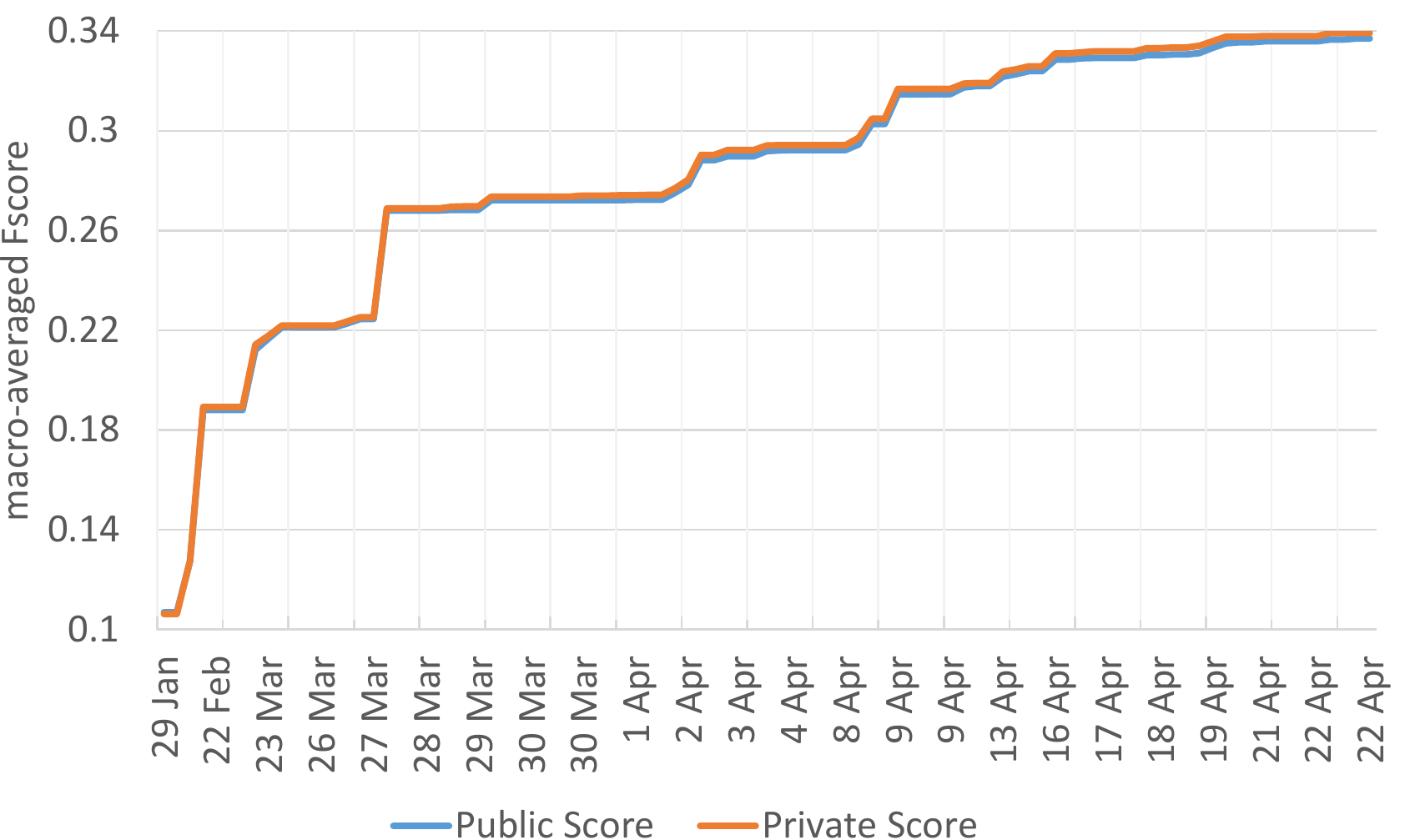}}
\caption{Progress on the test.txt macro-averaged Fscore during the competition. Growing the ensemble brought steady improvement, implementing the
transposed prediction caused jumps in the maFscore.}
\label{fig1}
\end{figure}

A second difficult complication in the contest was the choice of maFscore for evaluation measure, in contrast to earlier LSHTC competitions. What surprised the
contestants was that optimization of maFscore with high numbers of labels is problematic, since most labels will be missing. With maFscore a label occurring once
is just as important as one occurring 1000 times, and a label never predicted and one predicted by a 1000 false positives have the same effect on the score. 
Combined with most labels missing, normal optimization of classifiers proved difficult. It took us some time to figure out
the right way to solve this problem, but the solution made it possible for us to compete for the win. Before developing the transposed prediction used in both
the base-classifiers and the ensemble, our leaderboard score was around 22\%. A couple of simple corrections for maximizing maFscore correctly brought 
the ensemble combination close to 27\%, and using the transposed prediction with a larger and more diverse ensemble gave us the final score close to 34\%.
Other participants noticed this problem of optimizing maFscore, but likely most of them did not find a good solution.\\

The use of metafeature regression in the ensemble instead of majority voting provided a moderate improvement of about 0.5\%, and this much was needed
for the win. It is likely that the metafeatures optimized on the 23k comb\_dev.txt documents looked different from the metafeatures computed for the 452k
test.txt documents, even though the metafeatures were chosen or normalized to be stable to change in the number of documents. The optimal
amount of regularization for the Weka LinearRegression was untypically high at 1000. More complicated Weka regression models for the vote weight prediction 
failed to improve the test set score, likely due to overfitting the somewhat unreliable features. Another reason could be the small size of the comb\_dev.txt for 
ensemble combination. The ensemble fits the parameters for 55 metafeatures to predict the vote weight of each of the 42 base-classifiers, using only 23k points
of data shared by the 42 regression models. The improvement from Feature Weighted Linear Stacking could have been considerably larger, if a
larger training set had been segmented for the ensemble.\\

\section{Acknowledgements}
We'd like to thank Kaggle and the LSHTC organizers for their work in making the competition a success, and the machine learning group 
at the University of Waikato for the computers we used for our solution.
\bibliography{lshtc4}
\bibliographystyle{splncs}
\end{document}